\documentclass[11pt,a4paper]{article}

\usepackage{natbib}

\usepackage{amsmath}
\usepackage{amsfonts}
\usepackage{amsthm}
\usepackage{graphicx}
\usepackage[margin=3cm]{geometry}
\usepackage{multirow}
\usepackage{algorithmic}
\usepackage{algorithm}
\usepackage{subfigure}
\numberwithin{algorithm}{section}

\DeclareMathOperator{\tr}{tr}

\DeclareMathOperator{\diag}{diag}

\def\abovestrut#1{\rule[0in]{0in}{#1}\ignorespaces}
\def\belowstrut#1{\rule[-#1]{0in}{#1}\ignorespaces}
\def\abovespace{\abovestrut{0.20in}}

\def\belowspace{\belowstrut{0.10in}}

\usepackage{hyperref}
\usepackage{color}
\usepackage{xcolor}
\newtheorem{theorem}{Theorem}

\newcommand{\Acal}{{\cal A}}

\newcommand{\Ical}{{\cal I}}

\newcommand{\Pcal}{{\cal P}}

\newcommand{\Scal}{{\cal S}}

\newcommand{\Rmbb}{\mathbb{R}}

\begin{document} 

\title{Efficiently Using Second Order Information in Large $\ell_1$ Regularization Problems}

% unless you've provided the [accepted] option to the icml2013
\author{Xiaocheng Tang\thanks{Lehigh University, Bethlehem, PA 18015 USA. xct@lehigh.edu} \\ \and
Katya Scheinberg\thanks{Lehigh University, Bethlehem, PA 18015 USA. katyas@lehigh.edu} }
\maketitle
% \address{Lehigh University, Bethlehem, PA 18015 USA}

% % You may provide any keywords that you 
% % find helpful for describing your paper; these are used to populate 
% % the "keywords" metadata in the PDF but will not be shown in the document
% \keywords{optimization, large scale, graphical model, supervised learning, second-order information}
% 
% \vskip 0.3in

\begin{abstract} 
We propose a novel general algorithm LHAC that efficiently uses second-order information to train a class of large-scale $\ell_1$-regularized problems. Our method executes cheap iterations while achieving fast local convergence rate by exploiting the special structure of a low-rank matrix,  constructed via quasi-Newton approximation of the Hessian of the smooth loss function. A greedy active-set strategy, based on the largest violations in the dual constraints, is employed to maintain a working set that iteratively estimates the complement of the optimal active set. This  allows for smaller size of subproblems and eventually identifies the  optimal active set. Empirical comparisons confirm that LHAC is highly competitive with several recently proposed state-of-the-art specialized solvers for sparse logistic regression and sparse inverse covariance matrix selection.
\end{abstract}

\section{Introduction} % (fold)
\label{sec:introduction}

We consider convex sparse unconstrained  minimization problem of the following general form
\begin{align}
	\label{equ:origin}
	\min_{w}~F(w) = \lambda \|w\|_1 + L(w) 
\end{align}
where $L: \Rmbb^p \rightarrow \Rmbb$ is convex and twice differentiable and $\lambda > 0$ is the regularization parameter that  controls the sparsity of $w$. More generally, the regularization term $\lambda \|w\|_1$ can be extended to $\| \lambda \circ w \|_1 = \sum_{i=1}^p \lambda_i |w_i|$ to allow for different regularization weights on different entries, e.g., when there is a certain  sparsity pattern desired in the solution $w$. We will focus  on the simpler form as in (\ref{equ:origin}) in this work for the sake of simplicity of presentation, as the extension to the general form is straightforward. 

Problems of form (\ref{equ:origin}) have been the focus of much research lately in the fields of signal processing and machine learning. This form encompasses a variety of machine learning models, 
in which feature selection is desirable, such as sparse logistic regression \cite{Yuan2010,nGLMNET,shalev2009stochastic}, sparse inverse covariance selection \cite{Hsieh2011,Olsen2012,Sinco2009}, Lasso \cite{Tibshirani_1996}, etc. These settings often present common difficulties to optimization algorithms due to their large scale. During the past decade most optimization effort aimed at these problems focused on development of efficient first-order methods, such as accelerated proximal gradients methods \cite{Nesterov,Beck2009,Sparsa}, block coordinate descent methods \cite{nGLMNET,GLMNET,glasso_2008,Sinco2009} and alternating directions methods \cite{Alm_Scheinberg}. These methods enjoy low per-iteration complexity, but typically have low local convergence rates. Their performance is often hampered by small step sizes. This, of course, has been known about first-oder methods for a long time, however, due to the very large size of these problems, second order methods are often not a practical alternative.
In particular, constructing and storing a Hessian matrix, let alone inverting it, is prohibitively expensive for values of $p$ larger than $10000$, which often makes the use of the Hessian in large-scale problems impractical, regardless of the benefits of fast local convergence rate. 

Nevertheless, recently several new methods were proposed for sparse optimization which make careful use of second order information \cite{Hsieh2011,nGLMNET,Olsen2012,Chin2012}. These methods explore the following 
special properties of the sparse problems:  at the optimality many of the elements of $w$ are expected to equal $0$, hence methods which explore active set-like approaches can benefit from small sizes of subproblems. Whenever the subproblems are not small, these new methods exploit the idea that the subproblems do not need to be solved accurately. 
In particular we take a note of the following methods.

\citet{nGLMNET} proposes a specialized GLMNET \cite{GLMNET} implementation for sparse logistic regression, where coordinate descent method is applied to the lasso subproblem constructed using the Hessian of $L(w)$ -- the smooth component of the objective $F(w)$. Two major improvements are discussed to enhance GLMNET for larger problem -- exploitation of the special structure of the Hessian to reduce the complexity of each coordinate step so that it is linear on the number of training instances, and a two-level shrinking scheme proposed to focus the minimization on  smaller subproblems. \citet{Hsieh2011} later use the similar ideas in their specialized algorithm called QUIC for sparse inverse covariance selection. Benefiting from both its active-set strategy, which eventually converges to the optimal nonzero subspace, and its efficient use of the Hessian, QUIC behaves as Newton-like algorithms and is able to claim quadratic local convergence. Another related line of work begins with paper by \citet{Wright2012}, which proposes and analyzes an algorithm that is characterized by a two-phase minimization step for obtaining the improving direction, with one phase where a gradient descent step is taken towards minimizing the subproblem, and an enhanced phase where a Newton-like step is carried out in the nonzero subspace resulted from the first phase. Similar ideas are also explored in \citet{Olsen2012}, in which a class of orthant-based Newton method is proposes such that Newton-like algorithm is applied in an orthant face which lies in a reduced subspace and is identified by first taking a steepest descent step. A backtracking line search, however, has to be put in place afterwards to project the step back onto the orthant whenever the step overshoots.

The above mentioned methods share similar attributes. In particular, all of them  incorporate actual Hessian  either by confining the subproblem minimization to a smooth subspace \cite{Wright2012,Olsen2012}, or by using it along with coordinate descent methods \cite{nGLMNET,Hsieh2011}. Active-set strategy is another key element shared by these approaches, which facilitates the use of the Hessian, often requiring only the \emph{reduced Hessian} rather than the full one, and more importantly, help identify the optimal nonzero subspace and eventually achieve (with the Hessian) fast local convergence. Unfortunately, however, most of those active-set methods are only able to shrink the size of the subspace significantly when the current iterate is close enough to the optimality. Some algorithms are aware of this, e.g., \citet{Wright2012} gives up the Hessian and returns to first-order steps if the size of the subspace exceeds 500, QUIC uses a small number $\epsilon$ (set to a constant) to control the subspace size, etc. Hence, the efficient use of second order information in large problems is still a challenge. Of the aforementioned methods, the ones by \citet{nGLMNET} and \citet{Hsieh2011} produce the most satisfying results. But we note that both are specialized algorithms that heavily depends on the special structure present in the Hessians of the corresponding models. Hence their use will be limited.
% !!!!!!!!!!!!!Here list the main ideas behind QUIC, Nocedal et al methods, GLMNET and Steve Wrights paper (whichever he mentioned at MOPTA). !!!!!!!!!!!!!
% 
% !!!!!!!!!!!!!A bit more explanation about the main ideas of the algorithms.!!!!!!!!!!!!!
 
In this work we make use of similar ideas as in \citet{nGLMNET} and \citet{Hsieh2011}, but further improve upon these ideas to obtain efficient general schemes. In particular, we use a different working set selection strategy than those used in \cite{Hsieh2011}. We choose to select a working set by observing the largest violations in the dual constraints. Similar technique has been successfully used in optimization for many decades, for instance in Linear Programming \cite{sprint1992} and in SVM \cite{Scheinberg2006}. As in QUIC and GLMNET we optimize the Lasso subproblems  using coordinate descent,
but we estimate the Hessian using limited memory BFGS method \cite{NoceWrig06} because the low rank form of the Hessian estimates reduce the per iteration complexity of each coordinate descent step to a constant. 

% !!!!!!!!!!!!!A bit more details here, I think.....!!!!!!!!!!!!!

Our goal here, as mentioned above, is to achieve second-order type convergence rate while maintaining a comparable per iteration complexity with that of most first-order methods. Following the approach as in \citet{nGLMNET} and \citet{Hsieh2011}, we apply coordinate descent methods iteratively to the \emph{lasso} subproblems constructed at the current point.
The acceleration of subproblem minimization, therefore, depends on controlling either the number of coordinate descent steps or the complexity of each individual step. The contributions of our work are thus twofolds. First of all, we adaptively maintain a working set of coordinates with the largest optimality violations, such that the steps we take along those coordinates always provide the best objective function value improvement.  The greedy nature of this approach helps reduce the violation of optimality conditions rather aggressively in practice while effectively avoiding zero updates, and more importantly, extends/shrinks (depending on the initial point) the working set incrementally until it converges to the complement of the optimal active set. Secondly, we explore the use of the Hessian and Hessian approximations in the coordinate descent framework. We show that each coordinate descent step can be greatly accelerated by the use of a special form of limited-memory BFGS method \cite{NoceWrig06}. For example, in the case of sparse logistic regression, given the Hessian of logistic loss $L(w)$ as $B = C X^T D X$ where $ D \in \Rmbb^{N \times N} $ is a diagonal matrix and $X \in \Rmbb^{N \times p}$ is the data matrix, the best implementation so far can only reduce the complexity of each coordinate descent step to $O(N)$ flops \cite{nGLMNET}, while with the help of LBFGS which approximates the Hessian by a low rank matrix $B = \gamma I - Q\hat Q$ where $Q \in \Rmbb^{p \times 2m} $ and $\hat Q \in \Rmbb^{2m \times p}$, we are able to bring that complexity down to a constant time $O(m)$, depending on the limited-memory parameter $m$ which is often chosen between 5 and 20. The key observation here is that the Hessian approximation obtained by LBFGS is low rank unlike the true Hessian, and that can be exploited to expedite the computation of $(Bd)_i$, the main expense of every coordinate descent step, by letting $(Bd)_i = \gamma d_i - q_i^T \hat d$, where $q_i$ is the $i$-th row vector of the matrix $Q$ and $\hat d$ is cached and updated using one column of $\hat Q$ for each step.

The paper is organized as follows. In Section \ref{sec:outerprob} we explain how the subproblems are generated using LBFGS updates and working set selection.
In Section \ref{sec:innerprob} we explain how coordinate descent is applied to solve the subproblems with low rank Hessians. In Section \ref{sec:Experiments} we present computations results on
two instances of sparse logistic regression and five instances of sparse inverse covariance selection. The results demonstrate significant advantage of our approach compared to the other methods which inspired this work.

\section{Outer iterations}\label{sec:outerprob}
Based on a generalization of the sequential quadratic programming method for nonlinear optimization \cite{NoceWrig06,Tseng2009}, our approach iteratively constructs a piecewise quadratic model to be used in the step computation. At iteration $k$ the model is obtained by expanding the smooth component $L(w)$ around the current iterate $w_k$ and keeping the $\ell_1$ regularization term, as follows:
\begin{align}
	\label{equ:compute_d_outer}
	d_k = \arg\min_{d,B_k \succ 0} \{\nabla L^T_k d + d^T B_k d + \lambda\|w_k + d\|_1\}
\end{align}
where $B_k$ can be any positive definite matrix \cite{Tseng2009}. In particular, we note that both \citet{Hsieh2011} and \citet{nGLMNET} choose $B_k$ to be the Hessian of $L(w)$, in which case the objective function of (\ref{equ:compute_d_outer}) will be simply composed of the second-order Taylor expansion of $L(w)$ and the $\ell_1$ term $\lambda\|w_k + d\|_1$.

An active-set method maintains a set of indices $\Acal$ that iteratively estimates the optimal active set $\Acal^*$ which contains indices of zero entries in the optimal solution $w^*$ of (\ref{equ:origin})
\begin{align}
	\Acal^* = \{i \in \Pcal ~|~ (w^*)_i = 0\}
\end{align}
where $\Pcal = \{1,2,...,p\}$.
We use $\Acal_k$ to denote the set $\Acal$ at $k$-th iteration. For those coordinates $i \in \Acal_k$, we fix its corresponding entry in the current iteration $(w_k)_i$ such that it does not change from current iteration to the next $(w_k)_i = (w_{k+1})_i$. That is, equivalently, to say that the descent step along that coordinate stays zero $(d_k)_i = 0$. Adding the active-set strategy to our descent step computation (\ref{equ:compute_d_outer}), we thus obtain
\begin{align}
	\label{equ:compute_d_outer_Ak}
	\nonumber d_k = \arg\min_{d} &\{\nabla L^T_k d + d^T B_k d + \lambda\|w_k + d\|_1 \\
	& \mbox{s.t.}~ B_k \succ 0, d_i = 0, \forall i \in \Acal_k\}
\end{align}
Next we are going to discuss the particular choice we make in selecting the positive definite matrix $B_k$ and $\Acal_k$, or its complement $\Ical_k = \{i \in \Pcal ~|~ i \notin \Acal_k\}$, which we refer to in this paper as the working set.
\subsection{Low-Rank Hessian Approximation $B_k$} % (fold)
\label{sub:low_rank_hessian_approximation_b_k_}

 % !!!!!!!!!!!!!Start with a general outline and description of the algorithm, skipping details, but introducing notations, such as $B_k$, $x_k$, $A_k$, etc.!!!!!!!!!!!!!

We make use of Theorem \ref{theorem:LBFGS_compact} from \cite{NoceWrig06}, which gives a specific form of the low-rank Hessian estimates, which we denote by  $B_k$.  $B_k$ is essentially a low-rank approximation of the Hessian of $L(w)$ through the well-known limited-memory BFGS method, which allows  the capture of the curvature information to help achieve a faster local convergence.  

% !!!!!!!!!!!!!The derivations below should be shorter (just final forms should be given) and better explained - what it $s_i$, what is $t_i$? !!!!!!!!!!!
\begin{theorem}
	\label{theorem:LBFGS_compact}
	Let $B_0$ be symmetric and positive definite, and assume that the $k$ vector pairs $\{s_i,t_i\}_{i=0}^{k-1}$ satisfy $s_i^Tt_i > 0, s_i = w_{i+1} - w_i$ and $t_i = \nabla L_{i+1} - \nabla L_i $. Let $B_k$ be obtained by applying $k$ BFGS updates with these vector pairs to $B_0$, using the formula:
	\begin{align}
		B_{k} = B_{k-1} - \frac{B_{k-1} s_{k-1} s_{k-1}^T B_{k-1}}{s_{k-1}^T B_{k-1} s_{k-1}} + \frac{y_{k-1} y_{k-1}^T}{y_{k-1}^T s_{k-1}}
	\end{align}
	
	We then have that
	\begin{align}
		\label{equ:compact_update}
		B_k = B_0 -
		\begin{bmatrix}
			B_0S_k	&T_k
		\end{bmatrix}
		\begin{bmatrix}
			S_k^TB_0S_k &L_k\\
			L_k^T	&-D_k
		\end{bmatrix}^{-1}
		\begin{bmatrix}
			S_k^TB_0\\
			T_k^T
		\end{bmatrix},
	\end{align}
	where $S_k$ and $T_k$ are the $p \times k$ matrices defined by
	\begin{align}
		\label{equ:ST_update}
		S_k = [s_0,...,s_{k-1}], \qquad T_k = [t_0,...,t_{k-1}],
	\end{align} 
	while $L_k$ and $D_k$ are the $k \times k$ matrices
	\begin{align}
		(L_k)_{i,j} &= 
		\begin{cases}
			s_{i-1}^Tt_{j-1} \quad &\mbox{if }i > j \\
			0	\quad &\mbox{otherwise,}
		\end{cases} \\
		D_k &= \diag[s^T_0t_0,...,s_{k-1}^Tt_{k-1}].
	\end{align}
\end{theorem}
For large-scale problems, BFGS method is often used in the limited-memory setting, known as the L-BFGS method \cite{NoceWrig06}. Note that the matrices $S_k$ and $T_k$ are augmented by one column every iteration according to (\ref{equ:ST_update}), and the update (\ref{equ:compact_update}) will soon become inefficient if all the previous pairs $\{s_i, t_i\}$ are used. 

In the limited-memory setting, we maintain the set $\{s_i, t_i\}$ with the $m$ most recent correction pairs by removing the oldest pair and adding a newly generated pair. Hence when the number of iteration $k$ exceeds $m$, the representation of the matrices $S_k, T_k, L_k, D_k$ needs to be slightly modified to reflect the changing nature of $\{s_i, t_i\}$, and $S_k,T_k$ are maintained as the $ p \times m$ matrices. Also, $L_k$ and $D_k$ are now the $m \times m$ matrices.

In the L-BFGS algorithm, the basic matrix $B_0$ may vary from iteration to iteration. A popular choice in practice is to set the basic matrix at $k$-th iteration to $B^{(k)}_0 = \gamma_k I$, where  
\begin{align}
	\label{equ:LBFGS_gamma}
	\gamma_k = \frac{t^T_{k-1}t_{k-1}}{t^T_{k-1}s_{k-1}}
\end{align}
which is proved effective in ensuring that the search direction is well-scaled  so that less time is spent on line search. With this particular choice of $B_0^{(k)}$, we define $Q$ and $R$ to be 
\begin{align}
	Q &= 
	\begin{bmatrix}
		\gamma_kS_k	&T_k
	\end{bmatrix} \\
	R &= 
	\begin{bmatrix}
		\gamma_kS_k^TS_k &L_k\\
		L_k^T	&-D_k
	\end{bmatrix}^{-1}	
\end{align}
and the formula to update $B_k$ becomes
\begin{align}
	\label{equ:LBFGS}
	B_k &= \gamma_k I - QRQ^T \\
		&= \gamma_k I - Q\hat Q \quad \mbox{with  } \hat Q= RQ^T
\end{align}
where $\gamma_k$ is given by (\ref{equ:LBFGS_gamma}). Hence, the work to compute $B_k$ only requires the matrix $Q$ which is a collection of previous iterates and gradient differences, and $R^{-1}$ that is a $2m$ by $2m$ nonsingular (as long as $s_i^Tt_i > 0 ~\forall i$) matrix whose inverse is thus easy to compute given $m$ is a small constant.

% subsection low_rank_hessian_approximation_b_k_ (end)

\subsection{Greedy Active-set Selection $\mathcal{A}_k(\Ical_k)$} % (fold)
\label{sub:greedy_active_set_selection}
An obvious choice of $\Ical_k$ would be $\Ical_k = \{1,...,p\}$, taking into account all the coordinates in subproblem minimization. But as we said earlier, this can be inefficient because there will potentially be many non-progressive steps where $z$ in (\ref{equ:compute_z_sol}) ends up being zero. For example, if $d=0$, $(B_kd)_j$ turns into zero and (\ref{equ:compute_z_sol}) becomes
\begin{align}
	\label{equ:compute_z_simplified}
	z = 
	\begin{cases}
		\frac{(\nabla L_k)_j + \lambda}{-(B_k)_{jj}} & \text{ if } (\nabla L_k)_j + \lambda \leq (B_k)_{jj} (w_k)_j, \\
		\frac{(\nabla L_k)_j - \lambda}{-(B_k)_{jj}} & \text{ if } (\nabla L_k)_j - \lambda \geq (B_k)_{jj} (w_k)_j, \\
		-(w_k)_j & \text{ otherwise. } 
	\end{cases}
\end{align}
which is equal to zero if the $j$th entry of the subgradient $\partial F$, defined below, is zero
\begin{align}
	\label{equ:subgradient_F}
	(\partial F(w_k))_j = 
	\begin{cases}
		(\nabla L(w_k))_j + \lambda \quad &(w_k)_j > 0, \\
		(\nabla L(w_k))_j - \lambda \quad &(w_k)_j < 0, \\
		\max\{|\nabla L(w_k))_j|-\lambda,0\} \quad &(w_k)_j = 0
	\end{cases}
\end{align}

% !!!!!!!!!!!!! Shorten discussion below, mostly to explain our own rule and only mention the others !!!!!!!!!!!!!. 

% And we know that at optimality $\partial F(w^*)$ diminishes and that as we come closer to optimality there will be more and more entries in $\partial F$ turning into zero, thus more coordinate updates ending up being zero. 

Hence, the possibly worst choice of $\Ical_k$ would be letting $\Ical_k = \{i \in \Pcal~|~(\partial F(w_k))_i=0\}$, which will result in nothing but all zeros in coordinate descent update. 

% Another similarly bad choice of $\Ical_k$ is $\Ical_k = \{i~|~( w_k)_i=0\}$. For example, consider minimizing the simple function $F(w)  = w_1 + w_2 + 2|w_1| + 2|w_2|$ from the initial point $w_1 = 0, w_2 = 100$. Our initial working set thus will be $\Ical_1 = \{ 1\}$ and the algorithm will then choose $w_1$ to update, but we find that the coordinate step for $w_1$ computed in (\ref{equ:compute_z_simplified}) returns zero, so the algorithm will again choose $w_1$ to update and get $0$ again, thus never moving towards the optimal solution which is obviously at $w^*_1 = w^*_2 = 0$.

Let us now consider two potentially good choice of $\Ical_k$
\begin{align}
	\Ical^{(1)}_k = \{i \in \Pcal ~|~(\partial F(w_k))_i \neq 0\}  
\end{align}
used by \citet{nGLMNET} and \citet{Hsieh2011} and
\begin{align}
	\Ical^{(2)}_k = \{i \in \Pcal~|~ (w_k)_i \neq 0\}  
\end{align}
considered by \citet{Wright2012} and \citet{Olsen2012}. Note that $\Ical^{(1)}_k  \supseteq  \Ical^{(2)}_k$, because in practice $(\partial F(w_k))_i$ can only be exactly zero if $(w_k)_i = 0$ according to (\ref{equ:subgradient_F}), so we have that if $(w_k)_i \neq 0$ then $(\partial F(w_k))_i \neq 0$. Also note that the two sets will both converge to the optimal active set $\Ical^{(1)}_k  =  \Ical^{(2)}_k = \Ical^*$ when the correct non-zeros have been identified in $w_k$ and $w_k$ is close enough to the optimal $w^*$, because in which case the violations in the dual constraints of those zero entries in $w_k$ will be zero such that $\Acal^{(1)}_k  = \{i \in \Pcal~|~(\partial F(w_k))_i = 0\} = \Acal^{(2)}_k = \{i \in \Pcal~|~ (w_k)_i = 0\} $. The above two observations can also be understood by representing $\Ical^{(1)}_k$ as the union of two sets $\Ical^{(1a)}_k$ and $\Ical^{(1b)}_k$, defined by
\begin{align}
	\Ical^{(1a)}_k &= \{i \in \Pcal~|~(\partial F(w_k))_i \neq 0 \mbox{ and } (w_k)_i \neq 0\} \\
	\Ical^{(1b)}_k &= \{i \in \Pcal~|~(\partial F(w_k))_i \neq 0 \mbox{ and } (w_k)_i = 0\}
\end{align}
and we have that $\Ical^{(2)}_k = \Ical^{(1a)}_k$ in general, and that $\Ical^{(2)}_k = \Ical^{(1)}_k$ only when $\Ical^{(1b)}_k = \emptyset$.

Now let us introduce our rule to select the working set $\Ical_k$. Particularly, Let us use $\hat i := \Ical^{(1)}_k(i)$ to denote the coordinate $\hat i \in \Ical^{(1)}_k$ that has the $i$th largest violations $|(\partial F_k)_{\hat i}| $ in the dual constraints. We then choose the working set for $k$-th iteration by
\begin{align}
	\label{equ:greedy_Ik}
	\Ical_k = \Ical^{(2)}_k \cup \bigcup_{i=1}^{s_k} \hat i
\end{align} 
where $s_k$ is a small integer chosen as a fraction of $|\Ical^{(1)}_k|$. Note that our $\Ical_k$ is largely based on $\Ical^{(2)}_k$, whose size, as discussed above, is smaller than $\Ical^{(1)}_k$ as used by \citet{nGLMNET} and \citet{Hsieh2011}. This, of course, enables us to solve a smaller subproblem (\ref{equ:compute_d_outer_Ak}). However, we also note that choosing $\Ical_k = \Ical^{(2)}_k$ in our case is a bad idea because it does not allow zero elements of $w$ to become nonzero, so the method may fail to converge. To ensure convergence, we have to let every coordinate enter or leave our working set $\Ical_k$, which is one purpose of the union of the set $\bigcup_{i=1}^{s_k} \hat i$.

\subsection{Line Search and Convergence Analysis} % (fold)
\label{sub:line_search_and_convergence_analysis}
After the step $d_k$ is computed, a line search procedure needs to be employed in order for the convergence analysis to follow from the framework by \citet{Tseng2009}. In this work, we adapt the Armijo rule, choosing the step size $\alpha_k$ to be the largest element from $\{\beta^0, \beta^1, \beta^2, ... \}$ satisfying
\begin{align}
	\label{equ:line_search}
	F(w_k + \alpha_k d_k) \leq F(w_k) + \alpha_k \sigma \Delta_k
\end{align} 
where $0 < \beta < 1, 0 < \sigma < 1$, and $\Delta_k := \nabla L_k^T d_k + \lambda ||w_k + d_k||_1 - \lambda ||w_k||_1$. The convergence analysis from \citet{Tseng2009} also requires the positive definiteness of the matrix $B_k$. Since we obtain $B_k$ using LBFGS update, it is guaranteed to be positive definite as long as the product $s^T_i t_i > 0$. We note that when $L(w)$ in (\ref{equ:origin}) is strongly convex, then $s^T_i t_i > 0$ holds at any two points, otherwise if $L(w)$ might be non-convex, then damped BFGS updating needs to be employed \cite{NoceWrig06}. 

% subsection line_search_and_convergence_analysis (end)

\begin{algorithm}[htb]
	\caption{LHAC: \scriptsize {\normalsize L}OW RANK {\normalsize H}ESSIAN {\normalsize A}PPROXIMATION IN ACTIVE-SET {\normalsize C}OORDINATE DESCENT }
	\label{alg:LHAC}
	\begin{algorithmic}[1]
		\STATE Choose an initial iterate $w_0$, the LBFGS parameter $m$, the working set parameter $s$
		\STATE Set $S, T, D, L$ to empty matrix
		\STATE Set $w_1 \leftarrow w_0, \gamma_1 \leftarrow 1, Q \leftarrow 0, \hat Q \leftarrow 0$
		\STATE Set the iteration counter $k \leftarrow 1$
		\STATE Compute the gradients $\nabla L_k$ and $\partial F_k$
		\WHILE{ optimality test returns false  }
			\STATE Set $ d_k \gets 0 $ and $ \hat d \gets 0 $
			\STATE Compute the \emph{working} set $\Ical_k = \Ical^{(2)}_k \cup \bigcup_{i=1}^{s} \hat i$
			\FOR{ each coordinate $j \in \mathcal{I}_k$  }
				\STATE Compute $(B_k)_{jj} = \gamma_k - q_j^T\hat Q_j$
			\ENDFOR
			\FOR{ each coordinate $j \in \mathcal{I}_k$  }
				\STATE Compute $(B_kd)_j = \gamma_k d_j - q_j^T \hat d$
				% \STATE Compute $z = -c + \Scal(c-b/a, \lambda/a)$, with $a = (B_k)_{jj},b = (\nabla L_k)_j + (B_kd)_j$ and $c = (w_k)_j + (d)_j$
				\STATE Compute $z$ according to (\ref{equ:compute_z_sol})
				\STATE Update $ d_k \gets d_k + ze_j $, $\hat d \gets \hat d + z \hat Q_j$
			\ENDFOR
			\STATE Compute the step size $\alpha_k$ according to (\ref{equ:line_search})
			\STATE Set $w_{k+1} \gets w_k + \alpha_k d_k $
			\STATE Compute the gradients $\nabla L_{k+1}$ and $\partial F_{k+1}$
			\STATE Set $t_k \leftarrow \nabla L_{k+1} - \nabla L_{k}$ and $s_k \leftarrow w_{k+1} - w_{k}$
			\STATE Update $S,T,D,L$ according to Theorem \ref{theorem:LBFGS_compact}
			\STATE Compute $Q,\hat Q$ according to Theorem \ref{theorem:LBFGS_compact}
			% \IF{$k > m$}
			% 	\STATE Delete the first column of $S$,$T$ and $L$
			% 	\STATE Delete the first row of $L$ and $D$
			% \ENDIF
			% \STATE Set $S \gets \begin{bmatrix}S	&s_k\end{bmatrix}, T \gets \begin{bmatrix}T	&t_k\end{bmatrix}$
			% \IF{$k = 1$}
			% 	\STATE Set $L \gets 0$
			% \ELSE
			% 	\STATE Compute $l \gets s_k^T \cdot [t_1,...,t_{k-1}]$
			% 	\STATE Set $L \gets [L;l]$ and $L \gets \begin{bmatrix}L	&0\end{bmatrix}$
			% \ENDIF				
			% \STATE Set $D \gets [D;s_k^Tt_k]$, $\gamma \gets \frac{t^T_k t_k}{t^T_k s_k}$, $Q \gets \begin{bmatrix}\gamma S	&T\end{bmatrix}$
			% \STATE Compute $\hat Q \gets \begin{bmatrix}\gamma S^TS &L\\L^T	&-D\end{bmatrix}^{-1} \cdot Q^T$
			% \STATE Set $k \gets k + 1 $							
			\label{algstep:update_w}
		\ENDWHILE
	\end{algorithmic}
\end{algorithm}

\section{The inner problem}\label{sec:innerprob} % (fold)

\label{sec:the_algorithm}
At $k$-th iteration given the current iterate $w_k$, we apply coordinate descent method to the  piecewise quadratic subproblem (\ref{equ:compute_d_outer_Ak}) to obtain the direction $d_k$.
Suppose $j$th coordinate in $d$ is updated, hence $d' = d + ze_j$ ($e_j$ is the $j$-th vector of the identity). Then $z$ is obtained by solving the following one-dimensional problem
%\begin{align}
%	\nonumber\min_z~~ &L_k + \nabla L_k^T (d + ze_j) + \lambda ||w_k + d + ze_j||_1\\
%	&+ \frac{1}{2} (d + ze_j)^T B_k (d + ze_j) 
%\end{align}
%Omitting the terms not dependent on $z$ and utilizing the fact that $\nabla L_k^T e_j = (\nabla L_k)_j, e_j B_kd = (B_kd)_j $ and $e_j^TB_ke_j = (B_k)_{jj}$, we obtain a neater form
\begin{align}
	\label{equ:compute_z_prob}
	\nonumber\min_z  &\frac{1}{2}(B_k)_{jj} z^2 + ( (\nabla L_k)_j + (B_kd)_j  )z \\
	&+ \lambda | (w_k)_j + (d)_j + z |
\end{align}
which is well known to have the following closed-form solution that can be obtained by one soft-thresholding step \cite{Donoho92de-noisingby,Hsieh2011}:
\begin{align}
	\label{equ:compute_z_sol}
	z = -c + \Scal(c-b/a, \lambda/a)
	% \begin{cases}
	% 	\frac{(B_k d)_j + (\nabla L_k)_j + \lambda}{-(B_k)_{jj}} & \text{ if } (\nabla L_k)_j + (B_kd)_j + \lambda \\
	% 	&\leq (B_k)_{jj} [(w_k)_j + (d)_j], \\
	% 	\frac{(B_k d)_j + (\nabla L_k)_j - \lambda}{-(B_k)_{jj}} & \text{ if } (\nabla L_k)_j + (B_kd)_j - \lambda \\
	% 	&\geq (B_k)_{jj} [(w_k)_j + (d)_j], \\
	% 	-((w_k)_j + (d)_j) & \text{ otherwise. } 
	% \end{cases}
\end{align}
with $a,b,c$ chosen to be $a = (B_k)_{jj}, b = (\nabla L_k)_j + (B_kd)_j $ and $c = (w_k)_j + (d)_j$.

% In Algorithm \ref{alg:GCDM} we summarize the main steps we just discussed in terms of applying coordinate descent methods to solving the subproblem (\ref{equ:compute_d})

As  mentioned above, the special low-rank update of $B_k$ provides us an opportunity to accelerate the coordinate descent process. To see how, let us recall that $B_k = \gamma_k I - Q\hat Q$,
%is given by a product of two matrices 
%\begin{align}
%	\label{equ:compute_Bk}
%	B_k = \gamma_k I - Q\hat Q
%\end{align}
where $Q \in \Rmbb^{p \times 2m}$, $\hat Q \in \Rmbb^{2m \times p}$ and $m$ chosen as a small constant. Clearly we do not need to explicitly store or compute $B_k$. Instead, since $B_k$ is only used through $(B_k)_{ii}$ and $(B_kd)_i$ when applying soft-thresholding steps to updating each coordinate $i$, we can only store the diagonal elements of $B_k$ and compute $(B_kd)_i$ on the fly whenever it is needed. Specifically,
% \begin{align}
% 	(B_k)_{ii} = \gamma_k - q_i^TRq_i
% \end{align}
\begin{align}
	(B_k)_{ii} = \gamma_k - q_i^T\hat q_i
\end{align}
where $q_i$ is the $i$th row of the matrix $Q$ and $\hat q_i$ is the $i$th column vector of the matrix $\hat Q$. To compute $(B_kd)_i$, we maintain a $2m$ dimensional vector $\hat d := \hat Qd$, then
\begin{align}
	(B_kd)_i = \gamma_k d_i - q_i^T \hat d
\end{align}
which takes $O(2m)$ flops, instead of $O(p)$ if we multiply $d$ by the $i$th row of $B_k$. Notice though, that after taking the soft-thresholding step $\hat d$ has to be updated each time by $\hat d \gets \hat d + z_i \hat q_i$. This update  requires little effort given that $\hat q_i$ is a vector with $2m$ dimensions where $m$ is often chosen to be between $3$ and $20$. However, we need to use extra memory for caching $\hat Q$ and $\hat d$ which takes $O(2mp + 2m)$ space. With the other $O(2p + 2mp)$ space for storing the diagonal of $B_k$, $Q$ and $d$, altogether we need $O(4mp + 2p + 2m)$ space, which can be written as $O(4mp)$ when $p \gg m$.

% subsection subsection_name (end)

% section the_algorithm (end)

\section{Experiments} % (fold)
\label{sec:Experiments}
% 
% !!!!!!!! Consider including the ad-hoc analysis of the number of flops for each method, as you did in your talks !!!!!!!!!

\subsection{Sparse Logistic Regression} % (fold)
\label{sub:sparse_logistic_regression}
The objective function of sparse logistic regression is given by
\begin{align*}
%	\label{equ:logistic_regression}
	F(w) = \lambda ||w||_1 + \frac{1}{N} \sum_{n=1}^N \log(1 + \exp(-y_n \cdot w^Tx_n))
\end{align*}
where $L(w) = \frac{1}{N} \sum_{n=1}^N \log(1 + \exp(-y_n \cdot w^Tx_n))$ is the average logistic loss function and $ \{ ( x_n, y_n ) \}^N_{n=1} \in ( \Rmbb^p \times \{-1,1\} ) $ is the training set. The number of instances in the training set and the number of features are denoted by $N$ and $p$ respectively. Note that the evaluation of $F$ requires $O(pN)$ flops and  to compute the Hessian requires $O(Np^2)$ flops. Hence, we choose such training sets for our experiment that $N$ and $p$ are large enough to test the scalability of the algorithms and yet small enough to be able to run on a workstation.  

We test the algorithms on both sparse and dense data sets. The statistics of those data sets used in the experiments are summarized in Table~\ref{tab:l1log_results}. In particular, one data set, RCV1 \cite{rcv1v2}, is a text categorization test collection made available from Reuters Corpus Volume I (RCV1), an archive of over 800,000 manually categorized newswire stories. Each RCV1 document used in our experiment has been tokenized, stopworded, stemmed and represented by 47,236 features in a final form as weighted vectors. Another data set is GISETTE, originally a handwriting digit recognition problem from NIPS 2003 feature selection challenge. Its feature set of size 5000 has been constructed in order to discriminate between two confusable handwritten digits: the four and the nine.

We compare LHAC with the  first order method FISTA \cite{Beck2009}, and GLMNET \cite{GLMNET,nGLMNET} which uses Hessian information. GLMNET in particular, originally proposed by \citet{GLMNET} and later improved specifically for sparse logistic regression by \citet{nGLMNET} has been shown by \citet{Yuan2010} and \citet{nGLMNET} to be the state-of-the-art for sparse logistic regression (their experiments include algorithm such as OWL-QN \cite{Andrew_Gao_2007}, TRON \cite{Lin_Moré_1999}, SCD \cite{shalev2009stochastic}, BBR \cite{Genkin_Lewis_Madigan_2007}, etc.), so we compare LHAC with GLMNET. We implemented all three algorithms, for fair comparison, in MATLAB, and all the experiments were executed on AMD Opteron 2.0 GHz machines with 32G RAM and Linux OS.

\begin{figure}[ht]
	\centering
	\subfigure[ GISETTE dense. (\#Instances = 2000).]
	{
 	   \includegraphics[width=0.45\columnwidth]{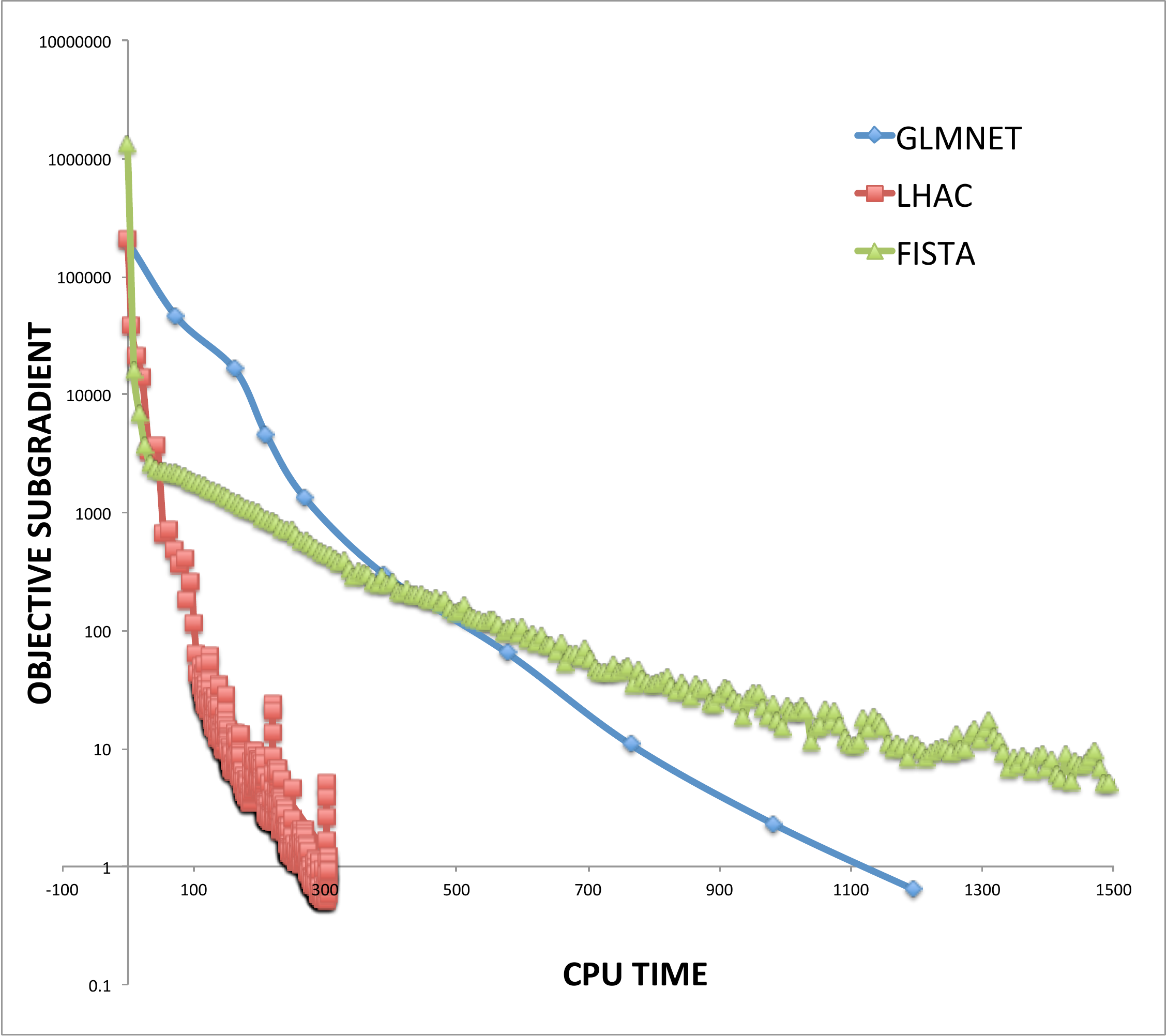}
	   \label{fig:gvt_git_2000}
	}
	~
	\subfigure[ RCV1 sparse. (\#Instances = 2500). ]
	{
 	   \includegraphics[width=0.45\columnwidth]{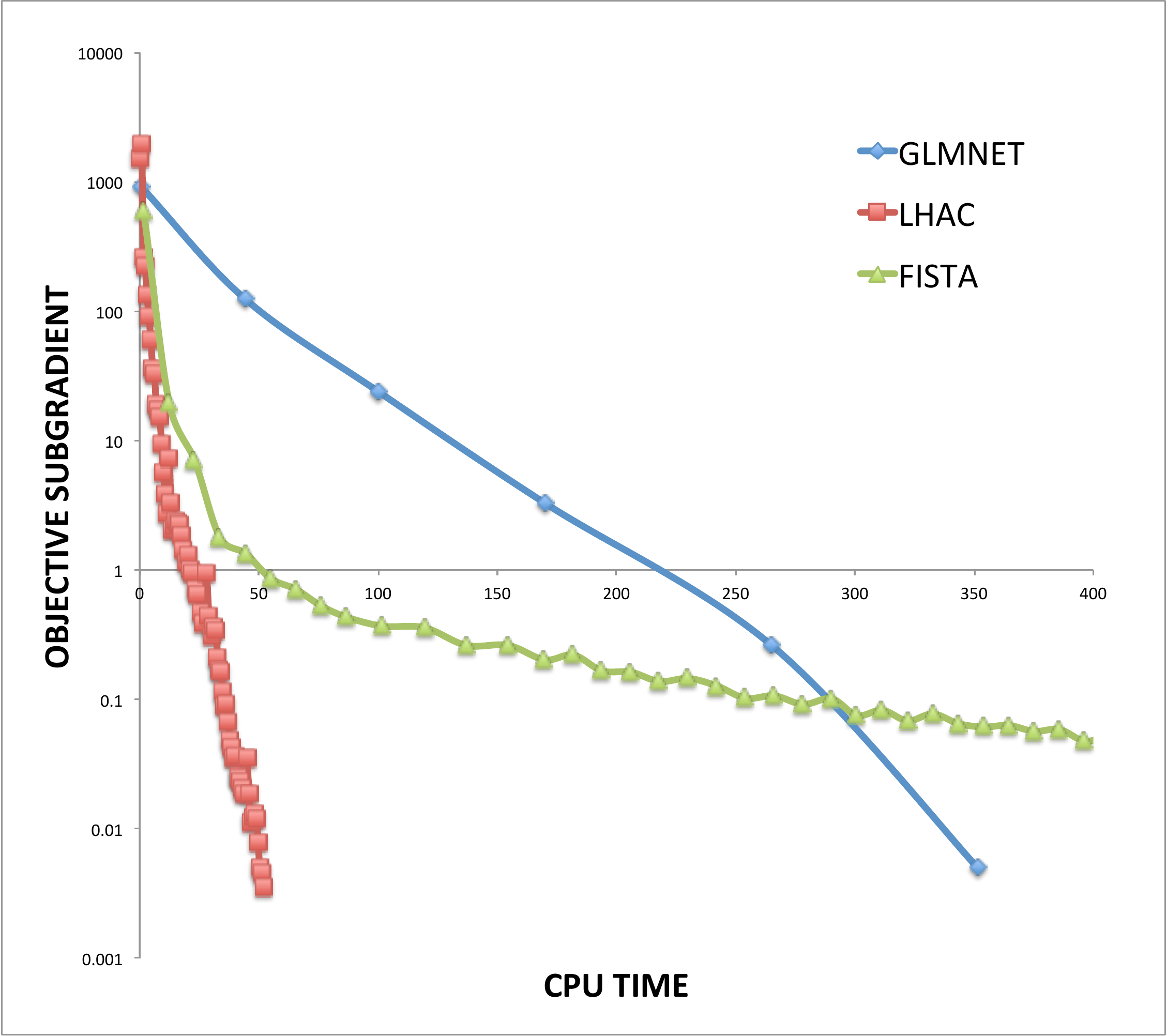}
	   \label{fig:gvt_rcv1_2500}
	}
	\\
	\subfigure[ GISETTE dense. \#Instances from 500 to 5000. ]
	{
 	   \includegraphics[width=0.45\columnwidth]{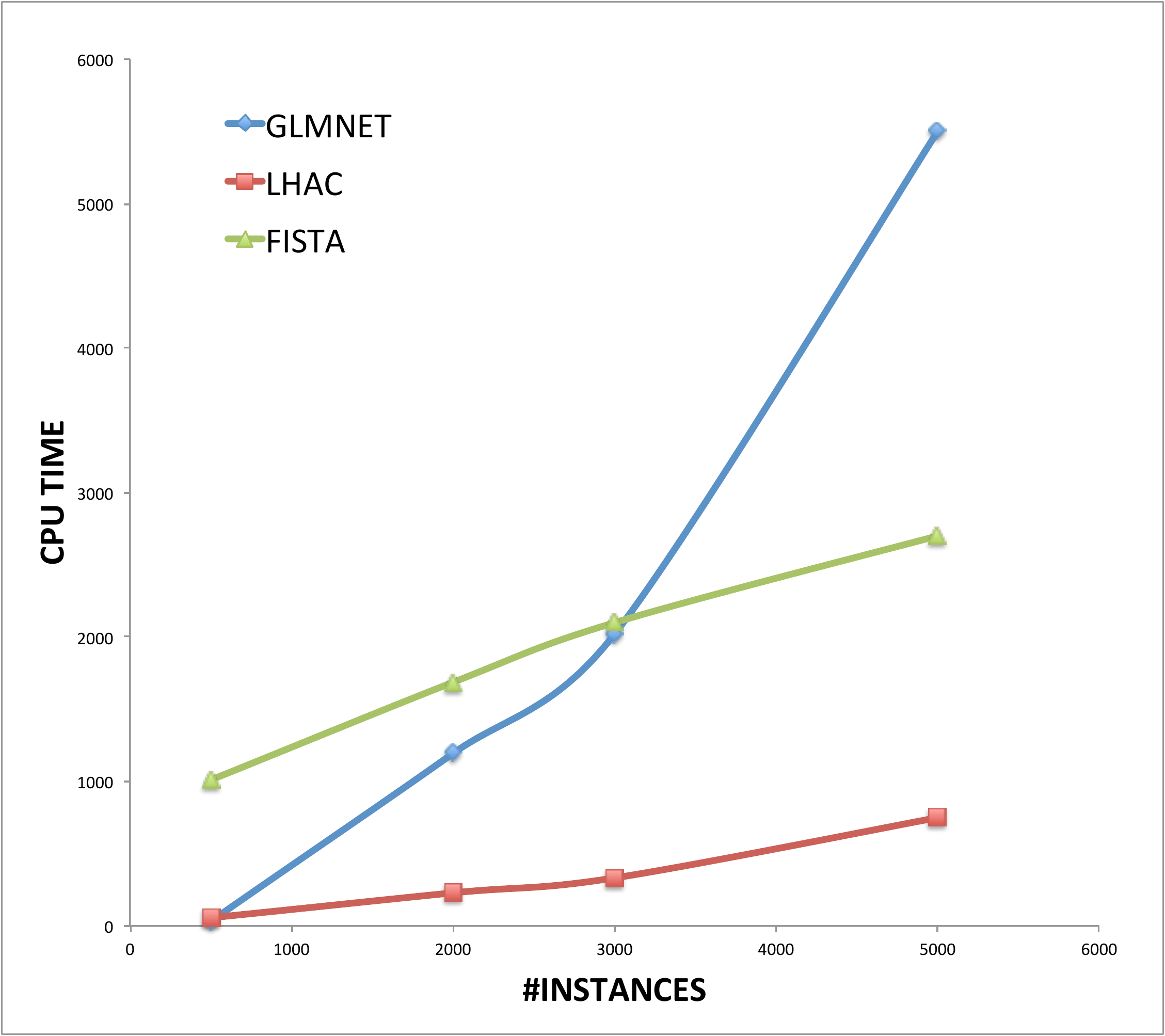}
	   \label{fig:Ivt_git}
	}
	~
	\subfigure[ RCV1 sparse. \#Instances from 500 to 2500.]
	{
 	   \includegraphics[width=0.45\columnwidth]{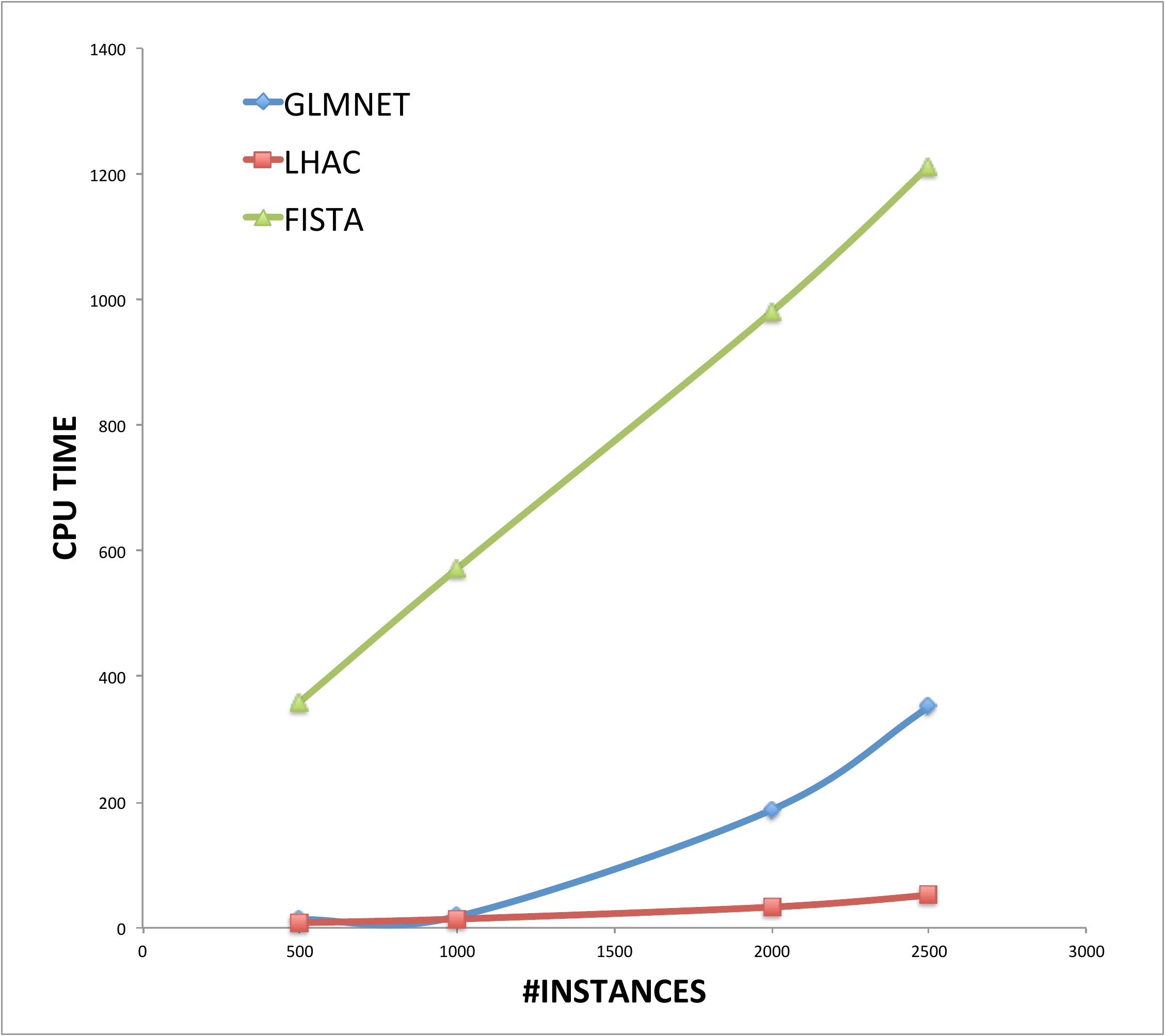}
	   \label{fig:Ivt_rcv1}
	}
		
	\caption{ Convergence plots in \ref{fig:gvt_git_2000} and \ref{fig:gvt_rcv1_2500} (the y-axes on log scale). Scalability plots in \ref{fig:Ivt_git} and \ref{fig:Ivt_rcv1}. }
	\label{fig:log_reg_exp}
\end{figure}

We have generated eight training sets, with different number of training instances, from the two data sets RCV1 and GISETTE, four for each one. For RCV1, the training size increases from 500 to 2500; for GISETTE, it ranges from 500 to 5000. In all the experiments, we terminate the algorithm when the current objective subgradient is detected to be less than the precision $\epsilon$ times the initial subgradient. For each training set, we solve the problem twice using each algorithm, one with a large epsilon, e.g., $\epsilon = 10^{-2}$, to test an algorithm's ability to quickly obtain a useful solution, the other one with a small epsilon, e.g., $\epsilon = 10^{-5}$, to see whether an algorithm can achieve an accurate solution. We report the runtime of each algorithm for all 16 experiments in Table~\ref{tab:l1log_results}. We also illustrate the results in Figure~\ref{fig:log_reg_exp}.

% \begin{figure}[ht]
% 	\centering
% 	\subfigure[dense dataset GISETTE (\#Instances = 2000).]
% 	{
%  	   \includegraphics[width=\columnwidth]{gvt_git_2000}
% 	   \label{fig:gvt_git_2000}
% 	}
% 	\subfigure[sparse dataset RCV1 (\#Instances = 2500). ]
% 	{
%  	   \includegraphics[width=\columnwidth]{gvt_rcv1_2500}
% 	   \label{fig:gvt_rcv1_2500}
% 	}	
% 		
% 	\caption{Plot of CPU time versus objective subgradient. $\epsilon$ is set to $ 10^{-5}$. }
% 	\label{fig:cpu_subgradient}
% \end{figure}

% \begin{figure}[ht]
% 	\centering
% 	\subfigure[sparse dataset RCV1. \#Instances from 500 to 2500.]
% 	{
%  	   \includegraphics[width=\columnwidth]{Ivt_rcv1}
% 	   \label{fig:Ivt_rcv1}
% 	}
% 	\subfigure[dense dataset GISETTE. \#Instances from 500 to 5000. ]
% 	{
%  	   \includegraphics[width=\columnwidth]{Ivt_git}
% 	   \label{fig:Ivt_git}
% 	}	
% 		
% 	\caption{Comparing the scalability of LHAC with that of GLMNET and FISTA; all three algorithm solve the problem to the same precision $\epsilon$ and we report the elapsed cputime versus the number of training instances. }
% 	\label{fig:cpu_samplesize}
% \end{figure}

As can be seen from the results, the advantage of LHAC really lies in the fact that it absorbs the benefits from both the first order and the second order methods, such that it iterates fast, with low per iteration cost, and that it converges fast like other quasi-Newton methods, by taking advantage of the objective curvature information. For example, let us look at how FISTA compares with GLMNET in Table~\ref{tab:l1log_results}, which provides a good case to study the well-known trade-offs between first order and second order methods. Particularly, it can be seen that in most cases when $\epsilon$ is set to $10^{-2}$, FISTA takes up much less time to terminate than GLMNET. However, FISTA falls short when high precision is demanded, e.g., $\epsilon$ set to $10^{-5}$, in which case GLMNET almost always terminates faster than FISTA. Notably, LHAC is able to perform well in both cases. In fact, it is overwhelmingly faster than the other two in all but one experiment (the runtime difference between LHAC and GLMNET is close when the training size is sufficiently small), which brings us to another interesting aspect about LHAC. That is, the larger the size of the training set is, the larger the margin becomes between LHAC and GLMNET. We illustrate this observation in Figure~\ref{fig:Ivt_git} and \ref{fig:Ivt_rcv1}, where the runtime of each algorithm ($\epsilon = 10^{-5}$) is plotted against the training size. Note that the complexity of LHAC scales almost linearly with the problem size  (with a much flatter rate than FISTA), while that of GLMNET increases nonlinearly. Figure~\ref{fig:gvt_git_2000} and \ref{fig:gvt_rcv1_2500} plot the change of objective subgradient against elapsed cputime in one experiment. Again, it can be observed that LHAC iterates as efficient as FISTA in the beginning, and while FISTA gradually slows down near the optimality, it continues to work as GLMNET until reaching the optimality tolerance $10^{-5}$.
% Another reason we choose FISTA and GLMNET to compare is that general framework... 

\begin{table}[ht]
\caption{CPU time comparisons on real world classification data sets. Data set $rcv1$ has $47,326$ features and $0.17\%$ non-zeros; Data set $gisette$ has $5,000$ features and $99\%$ non-zeros. $rcv1_{500}$ denotes that the training set is $rcv1$ and $500$ training samples are used.  $\epsilon$ indicates the optimization tolerance on the objective subgradient. The results show that LHAC is significantly faster than other methods.  }
\label{tab:l1log_results}
\vskip 0.15in
\begin{center}
\begin{small}
\begin{sc}
\begin{tabular}{|lc|ccc|}
\hline
\abovespace\belowspace

\multirow{2}{*}{Data set} &  \multirow{2}{*}{$\epsilon$} & \multicolumn{3}{|c|}{cputime(in seconds)}\\
						&	&  fista &glmnet &lhac\\
\hline
\abovespace
\multirow{2}{*}{$rcv1_{500}$}     &$10^{-2}$ &	4.54	&	6.39	&	\textbf{2.68} \\
					&$10^{-5}$ &	358.72	&	12.04	&	\textbf{8.09}\\
\hline
\abovespace						
\multirow{2}{*}{$rcv1_{1000}$}	&$10^{-2}$  &	14.28	&	6.18	&	\textbf{3.53}\\
						&$10^{-5}$ &	572.41	&	18.13	&	\textbf{14.13}\\
\hline
\abovespace									
\multirow{2}{*}{$rcv1_{2000}$} 	&$10^{-2}$ &	15.28	&	97.87	&	\textbf{6.13}\\
						&$10^{-5}$ &	980.72	&	188.48	&	\textbf{33.05}\\
\hline
\abovespace	
\multirow{2}{*}{$rcv1_{2500}$}	&$10^{-2}$ &	17.71	&	170.13	&	\textbf{9.95}\\
						&$10^{-5}$ &	1212.92	&	351.60	&	\textbf{52.00}\\
\hline
\abovespace	
\multirow{2}{*}{$gisette_{500}$}	&$10^{-2}$ &	52.91	&	10.93	&	\textbf{3.23}\\
						&$10^{-5}$ &	1009.66	&	\textbf{35.95}	&	54.90\\
\hline
\abovespace	
\multirow{2}{*}{$gisette_{2000}$}	&$10^{-2}$ &	85.41	&	269.26	&	\textbf{35.02}\\
						&$10^{-5}$ &	1686.69	&	1195.48	&	\textbf{229.67}\\
\hline
\abovespace	
\multirow{2}{*}{$gisette_{3000}$}		&$10^{-2}$ &	101.52	&	507.65	&	\textbf{40.17}\\
						&$10^{-5}$ &	2241.95	&	2021.40	&	\textbf{332.83}\\
\hline
\abovespace	
\multirow{2}{*}{$gisette_{5000}$}	&$10^{-2}$ &	\textbf{102.78}	&	1758.98	&	112.11\\
						&$10^{-5}$ &	2661.95	&	5935.72	&	\textbf{752.87}\\
\hline
\end{tabular}
\end{sc}
\end{small}
\end{center}
\vskip -0.1in
\end{table}

% subsection sparse_logistic_regression (end)

\subsection{Sparse Inverse Covariance Matrix Estimation} % (fold)
\label{sub:sparse_inverse_covariance_matrix_estimation}

In this section we compare our algorithm LHAC with QUIC by \citet{Hsieh2011}, a specialized solver that has been shown by \citet{Olsen2012} and \citet{Hsieh2011} to be the state of the art for sparse covariance selection problem defined below
\begin{align}
	\label{equ:sics_obj}
	\min_{X \succ 0} \quad f(X) = -\log \det X + \tr(SX) + \lambda ||X||_1
\end{align}
where the optimization variables are in a matrix form $X \in \Rmbb^{p \times p}$ that is required to be positive definite. 

\begin{figure}[ht]
	\vskip 0.2in
	\begin{center}
		\centerline{\includegraphics[width=0.7\columnwidth]{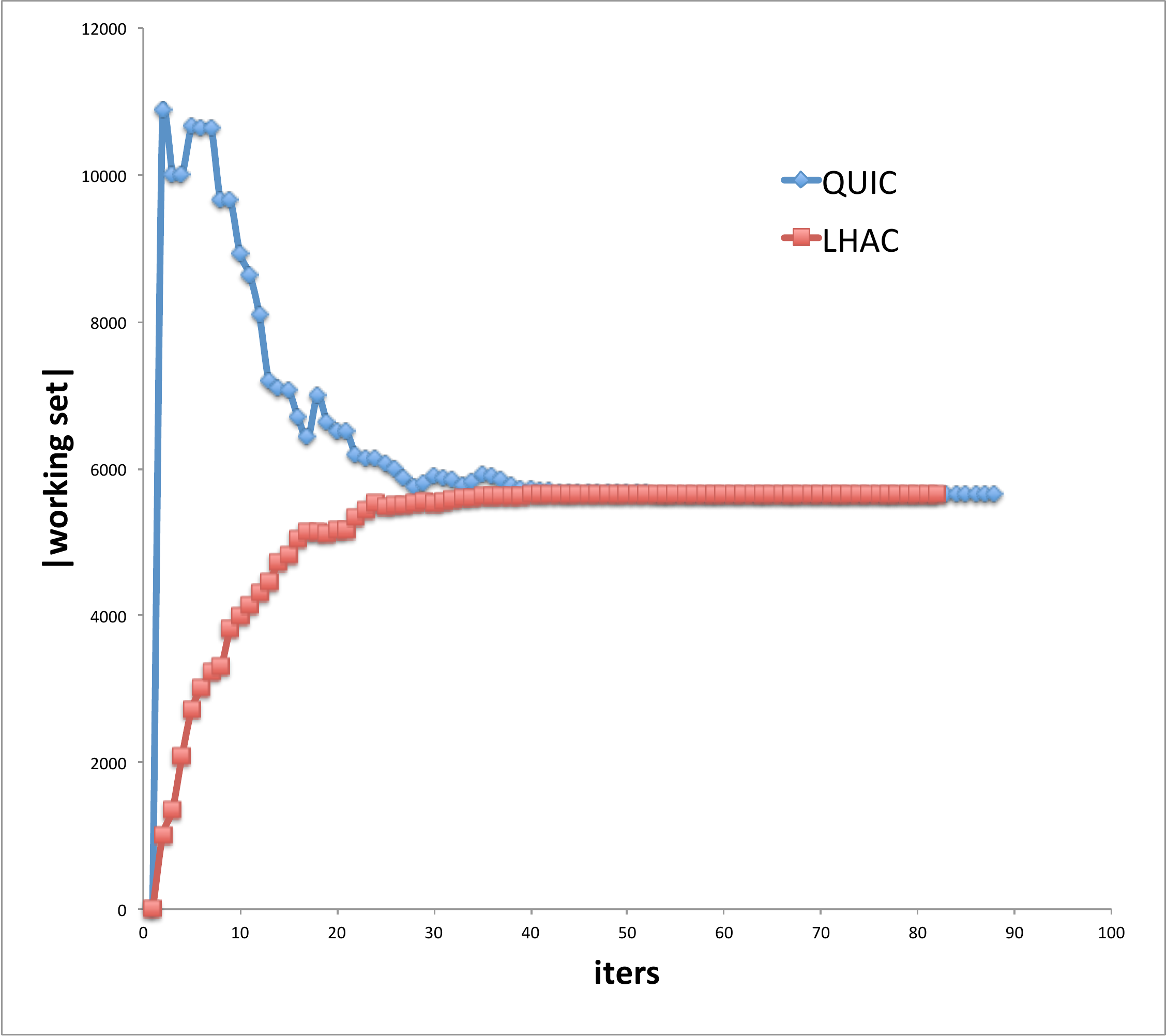}}
		\caption{Comparing two strategies - the strategy used by QUIC and the one used by LHAC - of choosing the working set on ER data set. Note that the working set maintained by LHAC approaches the optimal size from below while that of QUIC from above.  }
		\label{fig:work_set_size}
	\end{center}
	\vskip -0.2in	
\end{figure}

For this experiment, we are interested in comparing the total complexity of the two algorithms. CPU time, as often used, will not be a reasonable complexity measure in this case, because QUIC is implemented in C++ and LHAC - in MATLAB. Instead, we decide to count the number of flops required by each algorithm to solve problem (\ref{equ:sics_obj}). Here we note that the work required by both algorithm consists mainly two parts -- one for solving the subproblems and the other for computing Cholesky factorizations of $X$ during line search, and that the time either algorithm spends on the first part -- solving the subproblems -- generally takes around $95 \sim 99\%$ of the total elapsed time, as observed in the experiments. Hence we focus our comparisons on the first part of the work -- the one required by applying coordinate descent to solving subproblems, and we note that although LHAC generally requires more work in line search than QUIC (due to more outer iterations), it adds little to the total work for reasons stated above. 

\begin{table}[t]
\caption{Total complexity comparisons. $p$ stands for dimension. $\epsilon$ indicates the optimization tolerance on the objective subgradient.}
\label{tab:sics_results}
\vskip 0.15in
\begin{center}
\begin{small}
\begin{sc}
\begin{tabular}{|lcc|cc|}
\hline
\abovespace\belowspace

\multirow{2}{*}{Data set} & \multirow{2}{*}{$p$} & \multirow{2}{*}{$\epsilon$} & \multicolumn{2}{|c|}{Complexity($\times 10^6$) }\\
						  &						 &								&quic  &lhac\\
\hline
\abovespace
\multirow{2}{*}{Lymph}    & \multirow{2}{*}{587} 	&$10^{-2}$ &	47	&	\textbf{13} \\
						  &			&$10^{-6}$ &	105	&	\textbf{45}\\
\hline
\abovespace						
\multirow{2}{*}{ER}		&  \multirow{2}{*}{692}		&$10^{-2}$  &	218	&	\textbf{78}\\
						&			&$10^{-6}$ &	353	&	\textbf{217}\\
\hline
\abovespace									
\multirow{2}{*}{Arabidopsis}	&   \multirow{2}{*}{834}	&$10^{-2}$ &	559	&	\textbf{329}\\
								&		&$10^{-6}$ &	1001	&	\textbf{804}\\
\hline
\abovespace	
\multirow{2}{*}{Leukemia}	& \multirow{2}{*}{1255} 	&$10^{-2}$ &	2101	&	\textbf{574}\\
							& 	&$10^{-6}$ &	3028	&	\textbf{1326}\\
\hline
\abovespace	
\multirow{2}{*}{Hereditary}	& 	\multirow{2}{*}{1869}	&$10^{-2}$ 		&	16558			&	\textbf{8333}\\
							&				&$10^{-6}$ 	&	18519	&	\textbf{16770}\\
\hline
\end{tabular}
\end{sc}
\end{small}
\end{center}
\vskip -0.1in
\end{table}
 
Now let us describe how we compute the total complexity. Let us use $\kappa_o$ to denote the number of outer iteration and $\kappa_i$ to denote the number of inner coordinate sweep. Let $T_{CD}$ stand for the number of flops one coordinate descent step takes. Then we define the total complexity by
\begin{align}
	\label{equ:complexity}
	Complexity = \kappa_o \kappa_i p T_{CD}
\end{align}
In theory, the two algorithms have similar $\kappa_i$ because they both apply coordinate descent to a lasso subproblem, but different $\kappa_o$ and $T_{CD}$. Particularly, $\kappa_o$ of QUIC is smaller than that of LHAC whereas $T_{CD}$ of LHAC is smaller than that of QUIC. The reason is that QUIC uses the actual Hessian matrix of the smooth part $\tr(SX)-\log \det X$ rather than the low-rank Hessian approximation as in LHAC, which results in a different convergence rate ($\kappa_o$) and also a different complexity ($T_{CD}$) in computing the coordinate descent step (\ref{equ:compute_z_sol}). As we discussed earlier, the special structure of the low-rank matrix enables LHAC to accelerate that step to $O(m)$ flops with $m$ a constant number (chosen as 10 for this experiment). Whereas in the case of QUIC, one coordinate step takes problem-dependent complexity $O(p)$ where $p$ is the dimension of $X$, but it achieves quadratic convergence when close to the optimality. The above observations make it interesting to see in practice how the $Complexity$, defined in (\ref{equ:complexity}), will compare for the two algorithms since it is not  obvious which one is better through theoretical analysis alone.

The results are presented in Table~\ref{tab:sics_results}, and in Figure~\ref{fig:work_set_size} where we compare our active-set strategy with the one used by QUIC. The way we compute the $Complexity$ is by counting the number of flops in the coordinate descent step in each algorithm. For QUIC, we use their C++ implementation \cite{Hsieh2011} and add a counter directly in their code; for LHAC, we use our own MATLAB implementation. We report the results on real world data from gene expression networks preprocessed by \citet{Li2010}.
% and provided to us by Shiqian Ma. 
We set the regularization parameter $\lambda = 0.5$ for all the experiments as suggested in \citet{Li2010}. Similarly to the sparse logistic regression experiments, we solve each problem twice with different precision $\epsilon = 10^{-2}$ and $\epsilon = 10^{-6}$. Note that in all experiments QUIC consistently requires more flops to solve the problem than LHAC does. In one case where low precision is used it takes QUIC nearly three times more flops; even in the case QUIC is best at, where high precision is demanded, it can take QUIC 1.5 times more flops to achieve the same precision as LHAC did.

\section{Conclusions} % (fold)
\label{sec:conclusions}

We have presented a general algorithm LHAC for efficiently using second-order information in training large-scale $\ell_1$-regularized convex models. We tested the algorithm on two instances of sparse logistic regression and five instances of sparse inverse covariance selection, and found that the algorithm is faster (sometimes overwhelmingly) than other specialized solvers on both models. The efficiency gains are due to two factors: the exploitation of the special structure present in the low-rank Hessian approximation matrix, and the greedy active-set strategy, which correctly identifies the non-zero features of the optimal solution.
% section conclusions (end)

% \bibliographystyle{icml2013}
\bibliographystyle{abbrvnat}
\bibliography{New_bib}

% section appendix (end)

\end{document}